COMPLEXITY-OPTIMIZED SPARSE BAYESIAN LEARNING FOR SCALABLE CLASSIFICATION TASKS

# Complexity-Optimized Sparse Bayesian Learning for Scalable Classification Tasks

Jiahua Luo, Chi-Man Wong, Chi-Man Vong

***Abstract*—Sparse Bayesian Learning (SBL) constructs an extremely sparse probabilistic model with very competitive generalization. However, SBL needs to invert a big covariance matrix with complexity $O(M^3)$ ($M$: feature size) for updating the regularization priors, making it difficult for problems with high dimensional feature space or large data size. As it may easily suffer from the memory overflow issue in such problems. This paper addresses this issue with a newly proposed diagonal Quasi-Newton (DQN) method for SBL called DQN-SBL where the inversion of big covariance matrix is ignored so that the complexity is reduced to $O(M)$. The DQN-SBL is thoroughly evaluated for non-linear and linear classifications with various benchmarks of different sizes. Experimental results verify that DQN-SBL receives competitive generalization with a very sparse model and scales well to large-scale problems.**

## I. Introduction

Currently, research on deep learning methods is prevalent in the machine learning community. However, it requires rather large-scale data for training a deep model and lacks interpretability for the model structure. In many applications, such as antispam, fault diagnosis, on-board controls, robot devices and other Internet of Things (IoT) applications with small training data or limited computational resources (CPU or memory), a non-deep learning method with competitive accuracy and relatively compact model is preferred.

Traditional machine learning methods, including Logistic Regression (LR) [1-2], Support Vector Machine (SVM) [3-4], Sparse Bayesian models (SBL) [5-6], and their variants, have gained significant attention in recent years for problems with small training data. These sparse learning methods strike a balance between model sparsity and generalization.

LR with $\ell 1$-penalty is particularly suitable for high-dimensional feature space problems and is widely adopted in various industries. LR provides a sparse and interpretable model structure, along with good generalization through probabilistic outputs. However, it requires an exhaustive search for regularization parameters using cross-validation. SVM constructs classifiers in kernel space by maximizing the margin hyperplane, achieving sparsity with a minimal number of support vectors (samples). SVM offers state-of-the-art accuracy for small-data problems. However, the model size of SVM is typically proportional to the number of training data, resulting in a large model size without probabilistic predictions. Similar to LR, SVM relies on an exhaustive grid search of hyperparameters (penalty and kernel radius).

SBL is a probabilistic model for regression and classification. When modeling classification problems, SBL is built on the likelihood of training data. It can also be treated as a penalized LR model except that the regularization parameters in SBL are automatically determined by the training data, which avoids manual selection for hyperparameters. The regularization hyperparameters are assumed with Gaussian distributions, namely automatic relevance determination (ARD) prior [5], which provides competitive accuracy with a highly sparse model without significant deterioration of generalization. This advantage makes SBL very suitable and widely used for compressive sensing and recovery [7-10,28].

In many industrial devices with limited computational resources, the primary concern is not training time but rather fast execution time using a small yet highly precise model. SBL is well-suited for applications that require efficient memory usage and timely responsiveness. Within the SBL framework, several variants have been proposed, including the Relevance Vector Machine (RVM) [5] and the Sparse Bayesian Extreme Learning Machine (SBELM) [11-12], which offer competitive precision and sparsity. The RVM process is similar to that of SVM, where input features are mapped to a kernel space, but SBL is used for model training in RVM. On the other hand, SBELM is a type of MLP model in which features are randomly mapped to the hidden layer, and SBL is employed to select the most relevant hidden nodes, achieving sparsity. SBELM achieves comparable accuracy with shorter training time compared to RVM and is not affected by the size of the training data N, overcoming the bottleneck of RVM in modeling problems with large training instances.

An alternative methodology for solving the SBL is the Expectation-Maximization(EM) algorithm, in which the learning of SBL contains two main stages [5,13,14]:

(1) **MAP-Stage(M-step)**: Maximize the likelihood function (1) to estimate **w**, also interpreted as *maximum a posteriori* (MAP).

This work is funded by The Science and Technology Development Fund, Macau SAR (File No: 0112/2020/A). This work has been uploaded to arxiv https://arxiv.org/abs/2107.08195.

J. Luo is with the Department of Computer and Information Science, University of Macau, Macao SAR, China. (e-mail: luojiahuaguet@163.com).

C.-M. Vong is with the Department of Computer and Information Science, University of Macau, Macao SAR, China (e-mail: cmvong@um.edu.mo,).

C.-M. Wong is with the Department of Computer and Information Science, University of Macau, Macao SAR, China (e-mail: yc17417@um.edu.mo,).

$$\hat{\mathbf{w}} = \arg\max_{\mathbf{w}}[\ell(\mathbf{w}) + \log p(\mathbf{w})] \quad (1)$$

2) **A-Stage(E-step)**: Update the ARD prior $\boldsymbol{\alpha}$ using the diagonal elements of the covariance matrix (Hessian matrix inversion) calculated at the **MAP-Stage**.

Due to the large complexity of inverting the covariance matrix in SBL, which has a time complexity of $O(M^3)$($M$: feature dimension), it may encounter memory overflow issues, limiting its scalability for large problems. Specifically, RVM and its variants face a bottleneck in handling a large number N of training data due to the heavy complexity of $O(N^3)$. SBELM overcomes this limitation by mapping input features to hidden neurons. However, handling high-dimensional problems becomes challenging for SBELM, as it requires a significant number of neurons for effective generalization.

In the field of compressive sensing, some studies [7-10] effectively address this issue by treating it as a regression problem and utilizing SBL to estimate signal noise. They achieve this by selecting the most representative signals (i.e., a subset of relevant basis functions in RVM) to compress the original signals. Since the training data size is relatively small in this domain, these studies aim to reduce the complexity of covariance matrix inversion from $O(N^3)$ to $O(N^2)$ (where N represents the size of training data in RVM) by approximating a lower bound for the Maximum A Posteriori (MAP) estimation. Other studies [15-19] that employ SBL primarily focus on problems with small datasets. Some of these studies [17,19] adopt a greedy strategy to sequentially add or delete basis functions to/from the marginal likelihood maximization under the RVM framework. While these greedy sequential algorithms extend SBL to relatively large problems with a trade-off in accuracy and sparsity, yet they do not scale well for parallel or distributed computation modes in big data problems.

In recent years, Quasi-Newton methods such as LBFGS and Trust Region have achieved success in constructing logistic regression (LR) and support vector machines (SVM) for large-scale machine learning [1, 26, 27]. Intuitively, these methods seem appropriate for the MAP estimation in SBL. However, a critical issue in SBL is its reliance on the diagonal elements of the covariance inverse matrix (also known as the inverse Hessian matrix) obtained at the **MAP-stage** to update the Automatic Relevance Determination (ARD) prior. This inversion process incurs a computational complexity of $O(M^3)$ and requires $O(M^2)$ storage memory. This heavy computational burden becomes an obstacle for SBL to scale effectively for large problems. Consequently, most gradient-based solvers and Quasi-Newton methods that do not explicitly calculate the inverse Hessian matrix are not suitable for SBL, limiting the usage of classical SBL and its variants for large-scale machine learning problems.

In this paper, we propose a Diagonal Quasi-Newton (DQN) method, called DQN-SBL, for estimating the weight vector $\mathbf{w}$ and approximating the inverse Hessian matrix at the **MAP-stage** in SBL. In DQN-SBL, the covariance matrix inversion is ignored and instead approximated by a diagonal positive-definite matrix, while simultaneously estimating $\mathbf{w}$. Similar to the classical SBL, the diagonal elements are alternately employed to update the regularization priors. The contributions of this paper are summarized as follows:

1) Development of a diagonal Quasi-Newton (DQN) method to approximate the covariance matrix for the MAP function in SBL. This algorithm is both memory-efficient and computationally efficient, enabling effective handling of large-scale problems. Additionally, it can be easily implemented in parallel or distributed computation setups, further enhancing scalability.
2) The computational and storage complexity of DQN-SBL is linear with respect to the feature dimension $M$ (or $N$, the number of training data in RVMs). This reduction in complexity from $O(M^3)$ to $O(M)$ helps prevent memory overflow issues when dealing with large-scale problems.

The paper is organized as follows: Section II provides a brief review of SBL, outlining its key concepts and techniques. Section III introduces the proposed DQN-SBL method, which developing a Diagonal Quasi-Newton (DQN) approach to SBL. Section IV presents the experimental evaluations conducted to assess the performance of the proposed algorithms. Finally, Section V concludes the paper, summarizing the findings and discussing potential future directions for research.

## II. Short Review of SBL for Classification

### *A. SBL Linear Classification*

Given a set of $N$ training data $\boldsymbol{\mathcal{D}} = (\mathbf{x}_i,\ t_i),\ i = 1$ to $N$ with each $\mathbf{x}_i$ an $M$-dimensional vector and $t_i$ a binary output. In the SBL framework, the output $t$ is assumed with an independent Bernoulli distribution $p(t|\mathbf{x})$. Thus, for all training instances, the likelihood is the following probabilistic product

$$p(t|\mathbf{w},\mathbf{x}) = \prod_{i=1}^{N} \sigma\{\mathcal{Y}(\mathbf{x}_i;\mathbf{w})\}^{t_i}[1-\sigma\{\mathcal{Y}(\mathbf{x}_i;\mathbf{w})\}]^{1-t_i} \quad (2)$$

where the $\sigma(\cdot)$ is the sigmoid activation, $\sigma[\mathcal{Y}(\mathbf{x};\mathbf{w})] = 1/(1+e^{-\mathbf{x}^T\mathbf{w}})$. For the sake of simplicity, we denote $y = \sigma[\mathcal{Y}(\mathbf{x};\mathbf{w})]$ hereafter and $\mathbf{w} = (\mathrm{w}_0,\cdots,\mathrm{w}_M)^T$ the associated training weights. Besides, in SBL, each output weight $\mathrm{w}_k \in \mathbf{w}$ ($k = 0,\ldots,M$) is assumed with an independent zero-mean Gaussian distribution on an automatic relevance determination (ARD) hyperparameter $\alpha_k$ [5] such that

$$p(\mathrm{w}_k|\alpha_k) = \mathcal{N}(\mathrm{w}_k|0,{\alpha_k}^{-1}) \quad (3)$$

Then the distribution over $\mathbf{w}$ is the product of $p(\mathrm{w}_k|\alpha_k)$, which can be formulated as

$$p(\mathbf{w}|\boldsymbol{\alpha}) = \prod_{k=0}^{M} \frac{\alpha_k}{\sqrt{2\pi}} \exp(-\frac{\alpha_k \mathrm{w}_k^2}{2}) \quad (4)$$

Where $\boldsymbol{\alpha} = [\alpha_0,\cdots,\alpha_M]^T$. Every $\alpha_k$ is a positive value, the variance of $\mathrm{w}_k$ and governs its complexity. Each $\alpha_k$ can be further assumed with a Gamma distribution [5], whose parameters are usually set to zeros. The estimation of $\mathbf{w}$ in SBL is to approximate a Gaussian distribution over $\mathbf{w}$. Since this is an essential procedure for integrating the probabilistic marginal distribution for $p(\mathbf{t}|\boldsymbol{\alpha},\mathbf{X}) = \int p(\mathbf{t}|\mathbf{w},\mathbf{X})p(\mathbf{w}|\boldsymbol{\alpha})\, d\mathbf{w}$ by approximating a Gaussian

$$p(\mathbf{w}|\mathbf{t},\boldsymbol{\alpha},\mathbf{X}) = \frac{p(\mathbf{t}|\mathbf{w},\mathbf{X})p(\mathbf{w}|\boldsymbol{\alpha})}{p(t|\boldsymbol{\alpha},\mathbf{X})} \propto \mathcal{N}(\widehat{\mathbf{w}},\boldsymbol{\Sigma}) \quad (5)$$

The Gaussian approximation is achieved by a Taylor expansion on the logarithmic $\log\{p(\mathbf{t}|\mathbf{w},\mathbf{X})p(\mathbf{w}|\boldsymbol{\alpha})\}$. The covariance $\boldsymbol{\Sigma}$ and the Gaussian mode $\widehat{\mathbf{w}}$ are attained by setting $\nabla_{\mathrm{w}} \log\{p(\mathbf{t}|\mathbf{w},\mathbf{X})p(\mathbf{w}|\boldsymbol{\alpha})\}|_{\widehat{\mathrm{w}}} = 0$, that

$$\widehat{\mathbf{w}} = \boldsymbol{\beta}\boldsymbol{\Sigma}\mathbf{X}\mathbf{t} \quad (6)$$

$$\boldsymbol{\Sigma} = (\mathbf{X}^{\mathrm{T}}\boldsymbol{\beta}\mathbf{X} + \mathbf{A})^{-\mathbf{1}} \quad (7)$$

Where $y_i = \sigma[\mathcal{Y}(\mathbf{x}_i;\mathbf{w})]$ and $\mathbf{A}$ is a diagonal matrix $\mathbf{A}$ = diag($\boldsymbol{\alpha}$). $\boldsymbol{\beta}$ is a $N \times N$ diagonal matrix with element $\beta_{ii} = y_i(1-y_i)$, here $\mathbf{t} = [t_1, \ldots, t_N]$ denotes the vector of $N$ true outputs in the training dataset. After approximated by the Gaussian mode $\widehat{\mathbf{w}}$ and covariance matrix $\boldsymbol{\Sigma}$, the integral for $p(t|\boldsymbol{\alpha},\mathbf{x})$ becomes tractable. SBL estimates the unknown parameters $\boldsymbol{\alpha}$ by setting the $\nabla_\alpha \log p(t|\boldsymbol{\alpha},\mathbf{x}) = 0$.

$$\frac{\nabla_\alpha \log p(t|\boldsymbol{\alpha},\mathbf{x})}{\partial\alpha_k} = \frac{1}{2\alpha_k} - \frac{1}{2}\Sigma_{kk} - \frac{1}{2}\widehat{\mathrm{w}}_k^2 = 0$$

$$\Rightarrow \alpha_k^{new} = \frac{1-\alpha_k\Sigma_{kk}}{\widehat{\mathrm{w}}_k^2} \quad (8)$$

Using (6), (7) and (8), the output weights $\mathrm{w}_k$ and its associated hyperparameters $\alpha_k$ can be iteratively determined. Refer to the ARD [5-6], during the iteration, many $\alpha_k$ can be tuned to infinity such that their associated $\mathrm{w}_k$ becomes zero. Since $\alpha_k$ is inversely proportional to the weight $\widehat{\mathrm{w}}_k^2$ with (3) and (8), such weights are not updated in the subsequent iterations, which results in a highly sparse model.

### *B. RVM*

The RVM is a kernel-based, non-linear sparse Bayesian learning model that shares a similar functional form with SVM. It provides comparable accuracy to SVM for regression and classification tasks. One main difference is that RVM eliminates the process of exhaustive search for the regularization hyperparameter. In the context of a RVM classifier, a Gaussian kernel is utilized,

$$K(\mathbf{x}_{k1},\mathbf{x}_{k2}) = \exp\left(-\| \mathbf{x}_{k1} - \mathbf{x}_{k2} \| / \sigma^2\right) \quad (9)$$

where $\mathbf{x}_{k1}$ and $\mathbf{x}_{k2}$ represent two arbitrary instances from the training dataset. The original input $\mathbf{X}$ is converted to a basis matrix with shape $N \times N$. The basis matrix is subsequently solved using the SBL approach. However, a challenge arises when the size of N becomes sufficiently large, as it results in memory constraints during the computation of the inverse Hessian matrix in SBL, This calculation has a complexity of $O(N^3)$

### *C. SBELM*

SBELM is a sparse Bayesian random projection feedforward neural network that consists of three layers (input-hidden-output). It offers superior accuracy and an exceptionally sparse model. SBELM effectively resolves the limitation of kernel-based RVM when dealing with a large amount of training data. It achieves this by randomly encoding the raw feature space into a wide range of hidden nodes and utilizing the SBL framework to select the most optimal representative nodes, resulting to a highly sparse model.

The hidden layer output of SBELM with $L$ random neurons is represented by

$$h_l = f(\boldsymbol{\theta}^l\mathbf{x}) + b_l \quad (10)$$

where $\boldsymbol{\Theta} = [\,\boldsymbol{\theta}^1, \boldsymbol{\theta}^2, \cdots, \boldsymbol{\theta}^L]$ is randomly generated weights, with each $\boldsymbol{\theta}^l (l = 1 \ldots M)$ a $1 \times M$ dimensional random generated vector. The converted $L$-dimensional $\mathbf{h}$ with is fed into the SBL for further learning.

## III. THE PROPOSED DQN-SBL

### *A. A Diagonal QN method to SBL(DQN-SBL)*

Previously, SBL and its variants required the calculation of the inverse Hessian matrix $\boldsymbol{\Sigma}$, which involved using a second-order Newton-Raphson method for the MAP function [5,11,12]. However, this approach presents two issues. Firstly, the complexity of computing the inverse Hessian matrix is $O(M^3)$ in this methodology. This can lead to memory overflow problems when dealing with high-dimensional or large-scale datasets, and it cannot be easily parallelized for scenarios with extensive training data. Consequently, most SBL and its variants are only suitable for addressing small-scale problems.

In convex optimization, an alternative approach is to use Quasi-Newton methods to approximate the inverse Hessian matrix. The most well-known representatives in the literature are the BFGS and L-BFGS methods [20]. However, L-BFGS is not suitable for SBL since it lacks an explicit Hessian matrix, which is essential for updating the regularization parameter $\boldsymbol{\alpha}$ using equation (8).

In BFGS, $L(\mathbf{w})$ is approximated by a second-order Taylor expansion. For any $\mathbf{w}$, define $\boldsymbol{\delta}_m = \mathbf{w}_{m+1} - \mathbf{w}_m$, and $\boldsymbol{\gamma}_m = \nabla\boldsymbol{L}(\mathbf{w}_{m+1}) - \nabla\boldsymbol{L}(\mathbf{w}_m)$ at $m^{th}$ iteration, The Hessian matrix is approximated with a full positive definite $\boldsymbol{H}$ achieved by solving

$$\boldsymbol{H}_{m+1} = \underset{\mathbf{H}}{\operatorname{argmin}} \| \boldsymbol{H} - \boldsymbol{H}_m \|_{\mathbf{w}}$$

$$s.\,t.\ \boldsymbol{H} = \boldsymbol{H}^T \text{ and } \boldsymbol{H}\boldsymbol{\delta}_m = \boldsymbol{\gamma}_m$$

$\boldsymbol{H}$ is updated by the following formula

$$\boldsymbol{H}_{m+1} = \boldsymbol{H}_m + \frac{\boldsymbol{\gamma}_m\boldsymbol{\gamma}_m^T}{\boldsymbol{\gamma}_m^T\boldsymbol{\delta}_m} - \frac{\boldsymbol{H}_m\boldsymbol{\delta}_m\boldsymbol{\delta}_m^T\boldsymbol{H}_m}{\boldsymbol{\delta}_m^T\boldsymbol{H}_m\boldsymbol{\delta}_m} \quad (13)$$

However, it is computational expensive to approximate and store such a $M \times M$ full matrix $\boldsymbol{H}$ for high dimensional problems with computational and storage complexity $O(M^2)$.When $M$ amplifies in large problems, the storage and update of $\boldsymbol{B}$ become intractable.

In this subsection, we develop a diagonal Quasi-newton (DQN) method for approximating $\boldsymbol{\Sigma}$ for $L(\mathbf{w})$. In this methodology, $\boldsymbol{\Sigma}$ is approximated by a diagonal matrix $\boldsymbol{B}$ instead of using a full matrix. $\boldsymbol{B}$ shall satisfies the weak secant equation [21].

Derived from the BFGS formula (13), we here develop the diagonal updating type for the BFGS formula. Instead of approximating the Hessian matrix, we develop its inverse type $\boldsymbol{B}$. Due to the fact that,

$$\boldsymbol{B}_{m+1}^{-1} = \text{diag}[(\boldsymbol{B}_{m+1}^{(1)})^{-1}, (\boldsymbol{B}_{m+1}^{(2)})^{-1} \dots (\boldsymbol{B}_{m+1}^{(k)})^{-1}]$$

By (13), we have

$$\boldsymbol{B}_{m+1}^{-1} = \boldsymbol{B}_m^{-1} + \frac{\boldsymbol{\gamma}_m \boldsymbol{\gamma}_m^T}{\boldsymbol{\gamma}_m^T \boldsymbol{\delta}_m} - \frac{\boldsymbol{B}_m^{-1} \boldsymbol{\delta}_m \boldsymbol{\delta}_m^T \boldsymbol{B}_m^{-1}}{\boldsymbol{\delta}_m^T \boldsymbol{B}_m^{-1} \boldsymbol{\delta}_m},$$

Then the diagonal elements are updated by

$$\boldsymbol{B}_{m+1} = (\frac{\mathbf{1}}{\boldsymbol{B}_m} + \frac{(\boldsymbol{\gamma}_m)^2}{\boldsymbol{\gamma}_m^T \boldsymbol{\delta}_m} - \frac{(\boldsymbol{\delta}_m)^2}{(\boldsymbol{B}_m)^2 (\boldsymbol{\delta}_m^T \boldsymbol{B}_m^{-1} \boldsymbol{\delta}_m)})^{-1} \quad (14)$$

The diagonal elements in $\boldsymbol{B}_0$ are initialized with positive values to ensure $\boldsymbol{B}_0$ is *positive-definite*. By (14) the computational and storage complexity of $\boldsymbol{B}$ is reduced to $O(M)$.

As $\boldsymbol{B}$ in DQN is a diagonal matrix, when $L(\mathbf{w})$ converges to the minimum, the diagonal elements of $\boldsymbol{B}$ and $\mathbf{w}$ are repeatedly employed to update the ARD regularization parameters as

$$\alpha_k^{new} = \begin{cases} \frac{1-\alpha_k B_{kk}}{\widehat{w}_k^2}, \text{when } 1-\alpha_k B_{kk} \succ 0 \\ \frac{c}{\widehat{w}_k^2} \quad \text{otherwise} \end{cases} \quad (15)$$

During the iteration, there might exist a small subset of elements $1-\alpha_k B_{kk}$ tuned smaller than 0, which violates the distribution in (3). This occurs due to the occasional convergence of the approximation $\log p(t|\boldsymbol{\alpha}, \mathbf{x})$ occasionally converges to a local maximum[14]. To ensure the convergence of the iteration in such cases, we introduce a small constant $c$ as replacement of $1-\alpha_k B_{kk}$ in (15).

As it only contains element-wise floating computation in formula (15), whose complexity is $O(M)$. Using (14), the Storage and calculation of the full inverse Hessian matrix is ignored but replaced by a diagonal matrix $\boldsymbol{B}$ using $\boldsymbol{\delta}_m$ and $\boldsymbol{\gamma}_m$ in all iterations. The complexity is therefore reduced from $O(M^3)$ in the classical SBL to $O(M)$ in our algorithm, making it scale well to large-scale problems. We name this algorithm DQN-SBL in this paper.

### B. Summary of DQN

DQN is summarized in **Algorithm 1**. In our implementation, some settings are noted below.

*1)* $\boldsymbol{B}$ is initialized with an identity diagonal matrix and updated with Step 10 in **Algorithm 1**, in which $\boldsymbol{B}$ is guaranteed to be *positive definite* in the following iterations. Once the gradient $\nabla L(\mathbf{w}_m)$ falls below a given tolerance $\varepsilon$, a stable $\boldsymbol{B}$ is approximated. In order to guarantee convergence, a maximum number of iterations $MaxIts_qn$ is predefined.

*2)* The step size $\eta$ (Step 5, **Algorithm 1**) is normally determined by a line search algorithm that satisfies the wolf condition [20]. Variations of line search methods may result to different convergence speeds. In this paper, we do not describe the details about the line search methods [23]. The direction $\mathbf{p}_m$ here is scaled before imported to the *linesearch* method in step 5 to accelerate the learning speed. Once the convergence criterion is met, the approximated matrix $\boldsymbol{B}_m$ serves as the inverse Hessian matrix $\boldsymbol{\Sigma}$ and is used to update the hyperparameters $\boldsymbol{\alpha}$. This update is performed in the **A-stage** of the DQN-SBL algorithm. The specific details of how the hyperparameters $\boldsymbol{\alpha}$ are updated using $\boldsymbol{B}_m$ can be found in the implementation of the **Algorithm 2**.

**Algorithm 1** DQN method $[\mathbf{w}, \boldsymbol{B}] = \text{DQN}(\mathbf{X}, \mathbf{t}, \mathbf{w}, \boldsymbol{\alpha})$

Given objective function $L(\mathbf{w})$ and its gradient $\nabla L(\mathbf{w})$ function;
Initialize output weights vector $\mathbf{w}$;
Convergence tolerance $\varepsilon > 0$, maximum iteration $MaxIts_qn$;
1 iteration $m \leftarrow 0$
2 $\boldsymbol{B}_0 = diag(\mathbf{I})$ //O(M)
While $||\nabla l(\mathbf{w})|| > \varepsilon$ && $m < MaxIts_qn$
 3 $\mathbf{p}_m = -\boldsymbol{B}_m \cdot \nabla L(\mathbf{w}_m)$
 4 $\mathbf{p}_m = \mathbf{p}_m / ||\mathbf{p}_m||$ //acceralate convergence
 5 $\eta_m = linesearch(\mathbf{w}_m, \mathbf{p}_m)$
 6 $\boldsymbol{\delta}_m = \eta_m \mathbf{p}_m$
 7 $\mathbf{w}_{m+1} = \mathbf{w}_m + \boldsymbol{\delta}_m$
 8 $\boldsymbol{\gamma}_m = \nabla L(\mathbf{w}_{m+1}) - \nabla L(\mathbf{w}_m)$
 9 calculate $\boldsymbol{B}_m$ with (14)
 10 $m \leftarrow m+1$
12. return $\mathbf{w}_m$ and $\boldsymbol{B}_m$

### C. Summary of DQN-SBL

**Algorithm 2** summarizes the DQN-SBL algorithm for classification, which consists of two main stages: the **MAP-stage** for estimating output weights $\mathbf{w}$ and calculating the inverse Hessian matrix $\boldsymbol{B}$, and the **A-stage** for updating the regularization hyperparameters $\boldsymbol{\alpha}$. These stages are repeated until a termination condition is met.

In the **MAP-stage**, the weight vector w is estimated, and the inverse Hessian matrix $\boldsymbol{B}$ is computed. The step size $\eta$ is typically determined using the line search method, ensuring it satisfies the wolf-condition.

In the **A-stage**, the regularization parameters α are updated based on the converged $\widehat{\mathbf{w}}$ and $\boldsymbol{B}$ obtained from the **MAP-stage**. For certain $w_k$ values, if they tend to grow infinitely (exceeding a very large value), the associated weight component $w_k$ is pruned (set to zero), and these $\alpha_k$ values are not updated in subsequent iterations. Similar to SBL, once convergence is achieved in the **MAP-stage**, w and $\boldsymbol{B}$ are used to update $\boldsymbol{\alpha}$.

### D. Complexity analysis of DQN-SBL

In SBL, the largest complexity consumption is the calculation of the Hessian matrix inversion at the **MAP-Stage**, which takes $O(M^3)$ computational complexity and $O(M^2)$ storage memory. While in DQN-SBL, this inversion is replaced by (14). The memory for storages and calculation of $\boldsymbol{B}$, $\boldsymbol{\delta}$ and $\boldsymbol{\gamma}$ are $O(M)$.

Therefore, DQN-SBL scales well to high-dimensional problems (or large $N$ in RVMs).

**Algorithm 2:** Summary of DQN-SBL

**Initialization:**

Given dataset $\boldsymbol{\mathcal{D}} = (\mathbf{x}_i, \ \mathbf{t}_i), \ i = 1$ to $N$.

Given $MaxIts$, $ALPHA_MAX$, $DELTA_LOGALPHA$, a small *init_alpha, c*

$\mathbf{w} = [\mathbf{0}]_{L\times1}$, $\boldsymbol{\alpha} = init_alpha * [1]_{L\times1}$

**MAP-Stage:** conducting weights **w** and the inverse Hessian matrix $\boldsymbol{B}$

1. $[\mathbf{w}, \boldsymbol{B}] = \mathrm{DQN}(\mathbf{X}, \mathbf{t}, \mathbf{w}, \boldsymbol{\alpha})$

**A-Stage:** Estimation of hyperparameters $\boldsymbol{\alpha}$.

2. For every $\alpha_k$,
   Update $\alpha_k$ using (15)
   End for
3. For every $\alpha_k$ in $\boldsymbol{\alpha}$,
   If $\alpha_k > ALPHA_MAX$ //i.e., $\alpha_k \to \infty$
   prune $\alpha_k$ and the associated $w_k$;
   End if
   End for
4. If the largest absolute difference between two successive logarithmic values of $\alpha_k$ is lower than the *given* tolerance $DELTA_LOGALPHA$, then stop. Otherwise, repeat the **MAP-Stage** and **A-Stage** until the maximum iteration $MaxIts$.

## IV. Experimental Evaluation

### A. DQN-SBL for Non-linear Classification

In this section, we perform experiments to validate the effectiveness of DQN-SBL in non-linear classification by integrating into RVM and SBELM and comparing with their original versions.

Two sets of experiments are conducted: RVMs (DQN-RVM, RVM) and SBELMs (DQN-SBELM, SBELM). RVMs and SBELMs transform raw features into kernel space and random layers, respectively. The transformation method within each group of experiments remains the same. Here is a brief summary:

DQN-RVM: Integration of DQN-SBL into RVM.

DQN-SBELM: Integration of DQN-SBL into SBELM.

Initially, these four algorithms are evaluated on various benchmarks with relatively small datasets. Subsequently, DQN-SBL for linear classification is evaluated on several large-scale datasets with a significant amount of training data and high-dimensional feature space.

Since DQN-SBL and SBL were originally proposed for binary classifications, we employ the pairwise coupling (one-vs-one) method [11] to decompose multiclass classification into binary problems and then ensemble the results for the final prediction in our experiments.

#### 1) Experimental Setup

In our experiments, we utilized a total of 19 benchmarks. The properties of these datasets are provided in Table I. These datasets were obtained from the Libsvm machine learning repository [24], where they were sourced from other well-known machine learning repositories and preprocessed beforehand. For evaluation purposes, we selected the first 15 benchmarks, which have relatively small training data size or feature size, to assess the performance of DQN-RVM, RVM, DQN-SBELM, and SBELM. All features in these 15 datasets were linearly scaled to the range [-1, 1]. The remaining four benchmarks, characterized by large-scale discrete feature size and training data, were chosen to evaluate the linear DQN-SBL for classification.

Here are some of the settings used in our experiments:

a) In our implementation, we keep most settings the same for each group of experiments. Except that SBL is employed to RVM and SBELM, DQN-SBL to DQN-RVM and DQN-SBELM at **W-Stage** respectively. we set $DELTA_LOGALPHA = 10^{-3}$, $\text{ALPHA_MAX} = 10^{6}$, $MaxIts = 100$, $c$=0.0001 for **Algorithm 2** and $\varepsilon = 10^{-1}$, $MaxIts_qn = 100$ for **Algorithm 1**.

b) We implement the 4 algorithms based on the open-source toolboxes SBELM_v1.1 [25]. The activation function for DQN-SBELM, and SBELM for the hidden layer is sigmoid $\sigma(\mathbf{a}, b, \mathrm{x}) = 1/(1 + \exp(-(\mathbf{a}\mathrm{x} + b)))$, where $\mathbf{a}, b$ are the random generated synapses and bias with uniform distribution within $[-1, 1]$ respectively.

TABLE I
PROPERTIES OF TRAINING DATASET

| Dataset | Instances | Features | Classes |
|---|---|---|---|
| Adult1 | 1605 | 124 | 2 |
| colon | 62 | 2001 | 2 |
| breast | 683 | 11 | 2 |
| diabetes | 768 | 9 | 2 |
| german_numer | 1000 | 25 | 2 |
| a2a | 2265 | 121 | 2 |
| madelon | 600 | 502 | 2 |
| usps.binary | 2199 | 258 | 2 |
| Iris | 150 | 5 | 3 |
| Wine | 178 | 14 | 3 |
| segment | 2310 | 20 | 7 |
| satimage | 4435 | 37 | 6 |
| usps.all | 7291 | 258 | 10 |
| *news20* | 19304 | 1355191 | 2 |
| *real-sim* | 71362 | 20958 | 2 |
| *rcv1.binary.train* | 19663 | 47236 | 2 |
| *url_combined* | 515006 | 3231961 | 2 |

c) A 5-fold cross-validation strategy is used for evaluating the performance. We compare the mean validation accuracy, sparsity, and training time under the best hyperparameters. For RVMs, the kernel radius hyperparameter $\sigma$ is searched in $2^{\wedge[-5,5]}$.

d) For SBELMs, the number of hidden neurons and the seed $s$ for generating random hidden synapses and bias $[L, s]$ are taken from the grid search by [50, 100, 150, 200]× [1,2,3,4,5] for small feature size problems (<100), and [100, 500, 900, 1300]× [1,2,3,4,5] for others. In general, the accuracy for SBELM under different $s$ for generating random weights are with minor difference.

e) To collect more binary datasets, the benchmark *usps.binary* is fetched from the samples with label=1,2 from the *usps,* which originally contains 10 classes.

f) All experiments analyzed in this subsection are run on a 3.60GHz-CPU with 16-GB RAM PC.

TABLE II
COMPARISON OF ACCURACY

| Dataset | DQN-RVM | RVM | DQN-SBELM | SBELM |
|---|---|---|---|---|
| Adult1 | 75.39±0 | 75.39±0 | 75.39±0 | 75.39±0 |
| colon | 77.86±8.4 | 79.52±8.52 | 85.95±10.66 | 76.19±11.45 |
| breast | 96.94±1.47 | 97.35±1.26 | 96.91±1.35 | 96.78±0.97 |
| diabetes | 75.91±1.72 | 78.13±1.05 | 76.31±2.08 | 78.26±2.05 |
| german | 75.3±3.47 | 76.7±3.44 | 76.6±3.45 | 76.6±3.93 |
| a2a | 81.64±2.31 | 81.86±2.11 | 81.95±2.15 | 81.33±2.61 |
| madelon | 65.33±3.56 | 65±4.29 | 58±3.04 | 58.83±4.19 |
| *usps.binary* | 99.82±0.19 | 99.91±0.12 | 99.73±0.1 | 99.73±0.1 |
| Iris | 95.33±3.8 | 96.67±3.33 | 97.33±2.79 | 97.33±4.35 |
| Wine | 96.58±2.51 | 98.35±2.58 | 97.06±5.09 | 97.87±2.94 |
| segment | 94.55±0.1 | 96.62±0.86 | 95.76±0.85 | 96.97±0.59 |
| satimage | 89.49±1.38 | 90.78±1.17 | 86.45±1.83 | 88.86±1.26 |
| usps.all | 95.39±0.7 | 96.63±0.83 | 95.45±0.69 | 96.01±0.76 |

TABLE III
COMPARISON OF SPARSITY

| Dataset | DQN-RVM | RVM | DQN-SBELM | SBELM |
|---|---|---|---|---|
| Adult1 | 4.00 | 47.00 | 6.00 | 7.00 |
| colon | 6.60 | 7.00 | 22.00 | 9.00 |
| breast | 4.00 | 5.00 | 4.00 | 7.00 |
| diabetes | 125.00 | 8.00 | 4.00 | 10.00 |
| german | 190.20 | 11.00 | 10.00 | 15.00 |
| a2a | 383.00 | 20.00 | 47.00 | 94.00 |
| madelon | 480.00 | 480.00 | 49.00 | 85.00 |
| *usps.*binary | 8.40 | 5.00 | 7.00 | 7.00 |
| Iris | $3.67 \times 3$ | $4.00 \times 3$ | $3.67 \times 3$ | $4.00 \times 3$ |
| Wine | $3.67 \times 3$ | $5.33 \times 3$ | $3.00 \times 3$ | $4.33 \times 3$ |
| segment | $32.81 \times 21$ | $5.43 \times 21$ | $4.43 \times 21$ | $5.38 \times 21$ |
| satimage | $131.80 \times 15$ | $15.00 \times 15$ | $6.67 \times 15$ | $12.73 \times 15$ |
| usps.all | $119.69 \times 45$ | $8.25 \times 45$ | $45.24 \times 45$ | $51.37 \times 45$ |

We roughly measure the sparsity for the trained model using the number of the remaining basis functions and hidden nodes in RVMs and SBELMs.
$\times$ denotes multiplying the number of binary classifiers using the one-vs-one training strategy

TABLE IV
COMPARISON OF TRAINING TIME(SECONDS)

| Dataset | DQN-RVM | RVM | DQN-SBELM | SBELM |
|---|---|---|---|---|
| Adult1 | 1.88E+03 | 5.62E+02 | 4.84E+02 | 2.51E+02 |
| colon | 8.03E+01 | 1.08E+01 | 3.30E+02 | 1.48E+02 |
| breast | 5.38E+02 | 2.30E+02 | 8.86E+01 | 6.77E+00 |
| diabetes | 4.30E+02 | 1.23E+02 | 5.05E+01 | 5.58E+00 |
| german | 7.44E+02 | 2.21E+02 | 1.34E+02 | 2.03E+02 |
| a2a | 4.38E+03 | 1.29E+03 | 1.34E+03 | 3.31E+02 |
| madelon | 2.06E+02 | 3.01E+02 | 3.23E+02 | 1.53E+02 |
| *usps*.binary | 6.32E+03 | 3.14E+03 | 3.93E+02 | 1.63E+01 |
| Iris | 8.10E+02 | 5.34E+01 | 2.37E+02 | 7.14E+00 |
| Wine | 2.31E+02 | 5.99E+01 | 2.34E+02 | 1.46E+01 |
| segment | 4.36E+04 | 2.87E+03 | 1.62E+03 | 1.20E+02 |
| satimage | 2.44E+04 | 1.75E+04 | 2.06E+03 | 1.87E+02 |
| usps.all | 8.63E+04 | 2.16E+04 | 4.20E+04 | 2.05E+04 |

TABLE V
COMPARISON OF COMPLEXITIES

| Dataset | DQN-RVM / RVM (Ratio) | | DQN-SBELM / SBELM (Ratio) | |
|---|---|---|---|---|
| | Computational Complexity | Storage Complexity | Computational Complexity | Storage Complexity |
| Adult1 | **3.88E-07** | **6.23E-04** | **5.92E-07** | **7.69E-04** |
| colon | **2.60E-04** | **1.61E-02** | **5.92E-07** | **7.69E-04** |
| breast | **2.14E-06** | **1.46E-03** | **2.50E-05** | **5.00E-03** |
| diabetes | **1.70E-06** | **1.30E-03** | **2.50E-05** | **5.00E-03** |
| german | **1.00E-06** | **1.00E-03** | **2.50E-05** | **5.00E-03** |
| a2a | **1.95E-07** | **4.42E-04** | **5.92E-07** | **7.69E-04** |
| madelon | **2.78E-06** | **1.67E-03** | **5.92E-07** | **7.69E-04** |
| *usps*.binary | **2.07E-07** | **4.55E-04** | **5.92E-07** | **7.69E-04** |
| Iris | **4.44E-05** | **6.67E-03** | **2.50E-05** | **5.00E-03** |
| Wine | **3.16E-05** | **5.62E-03** | **2.50E-05** | **5.00E-03** |
| segment | **1.87E-07** | **4.33E-04** | **2.50E-05** | **5.00E-03** |
| satimage | **5.08E-08** | **2.25E-04** | **2.50E-05** | **5.00E-03** |
| usps.all | **1.88E-08** | **1.37E-04** | **5.92E-07** | **7.69E-04** |

*2) Comparison of Accuracy*

The comparison of the best accuracies achieved on these benchmarks is presented in Table II. The marginal decisive probabilistic score for all classifiers is set to 0.5. Overall, SBELMs (DQN-SBELM, SBELM) perform slightly better than RVMs (DQN-RVM, RVM). Let's analyze them separately. The accuracies of RVMs and SBELMs are very close in most benchmarks, except for colon where DQN-SBELM outperforms the others significantly (especially for feature sizes $> 3000$ and $> 7000$). This suggests that DQN-SBL is more robust when dealing with high-dimensional problems.

To compare the stability under different parameter settings, we depict the accuracy under different hyperparameters ($\sigma$) for RVMs and SBELMs in Figures 1 and 2, respectively. The accuracy curves for DQN-RVM and RVM are very similar in most benchmarks under different $\sigma$ values, with only small fluctuations observed in colon-2000, diabetes, and madelon. However, both models achieve similar best accuracies within the parameter selection range for these three benchmarks. On the other hand, the accuracy curves for SBELMs are much more stable under different L values, as shown in Figure 2. Nevertheless, SBELMs achieve competitive accuracy across most datasets. This validates the objective of our work: *achieving the same accuracy and sparsity while reducing computational and storage complexity.*

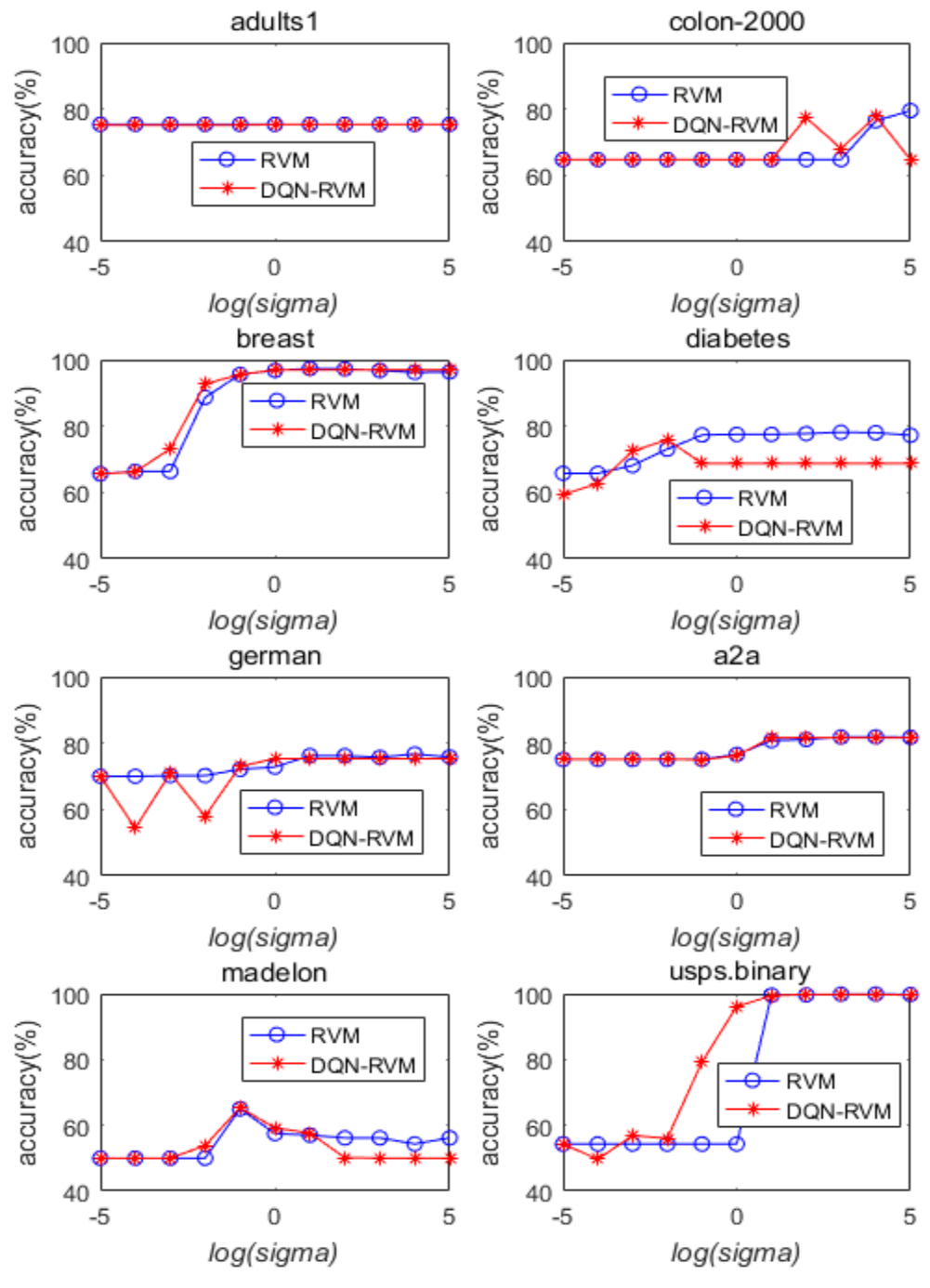

Fig. 1. Accuracy of RVMs under different hyperparameter ($log\ \sigma$)

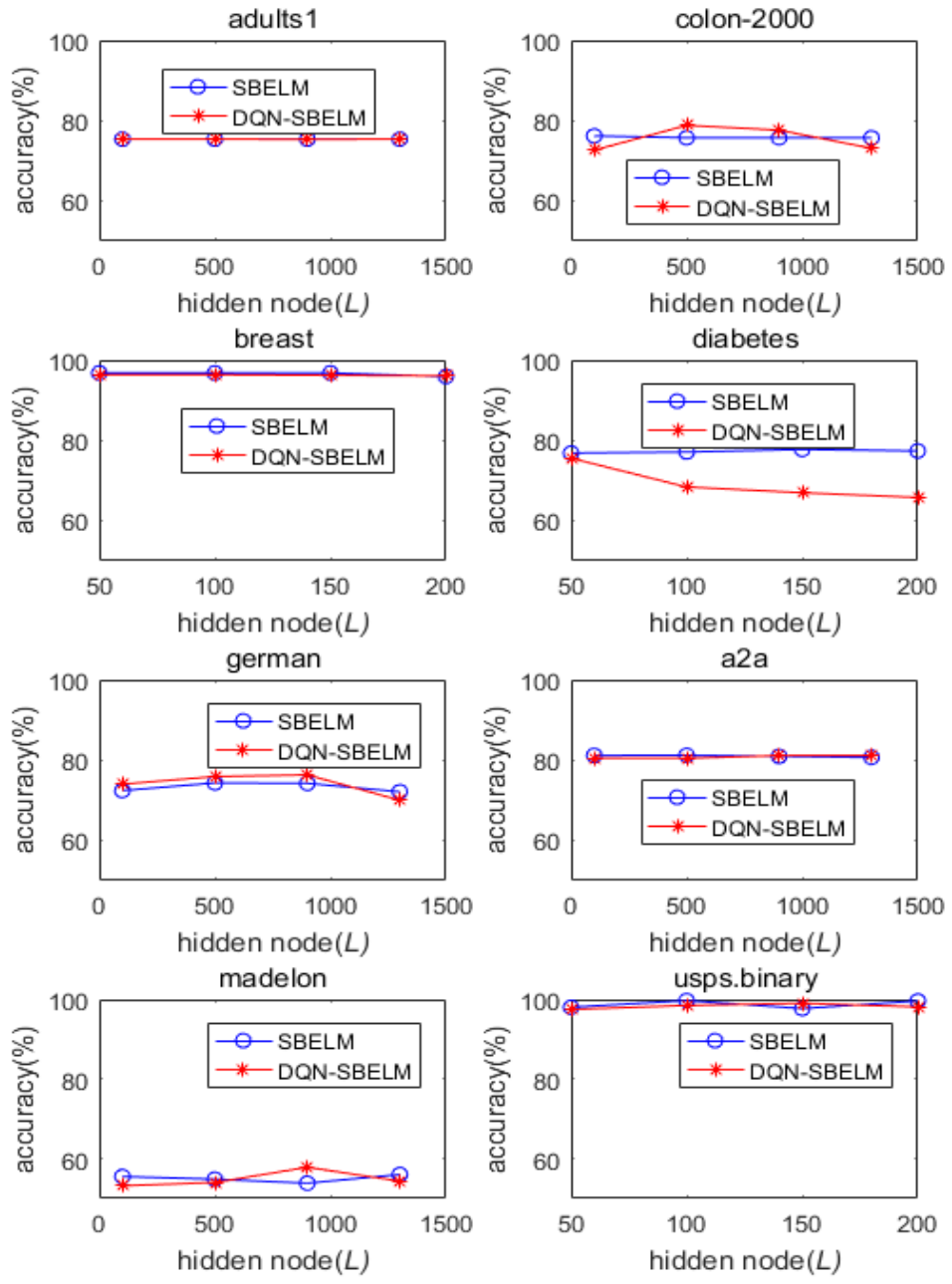

Fig. 2. Accuracy of SBELMs under different number of hidden nodes($L$).

*3) Comparison of Sparsity*

We assessed sparsity by measuring the average size of the remaining basis functions for RVMs and the remaining hidden nodes ($L_{remain}$) for SBELMs. Table III presents the sparsity values corresponding to the highest accuracy recorded in Table II. Overall, the sparsity in SBELMs is of a similar magnitude. The sparsity values for RVMs are very close, although DQN-RVM exhibits some fluctuations for certain benchmarks. By referring to Fig. 1, it is evident that RVMs perform better with larger $\sigma$ values, resulting in smaller sparsity and improved accuracies. However, the sparsity of DQN-RVM is significantly smaller than the training data size, yet its corresponding accuracy in Table II is comparable to the original RVM. In practical applications involving large-scale training data, RVM may not be the preferred choice, but DQN-RVM effectively addresses this challenge.

Furthermore, we present the sparsity analysis under different σ and L for RVMs and SBELMs on the first 8 datasets in Figure 3 and Figure 4, respectively. In general, RVM exhibits higher sparsity than DQN-RVM at smaller $\sigma$ values in Figure 3. This can be attributed to the presence of a small subset of $1-\alpha_k B_{kk} < 0$ during training, where the truncation of some small $|w_k|$values is prevented due to the update of $\alpha_k^{new}$ by $\frac{c}{\hat{w}_k^2}$. However, this does not significantly affect the accuracy and remains close to the RVM algorithm, as depicted in Figure 1. This phenomenon becomes less prominent as $\sigma$ increases, where the best accuracy is typically achieved, thus confirming our previous analysis in Section III.B.

In Figure 4, the sparsity of DQN-SBELM is lower than that of SBELM across different L values, and it remains stable as $L$ increases for both DQN-SBELM and SBELM. The accuracy of these two models remains consistently stable under varying $L$ values.

The sparsity of SBELMs remains consistently close and stable across most datasets. However, when comparing the DQN-based models, DQN-SBELM achieves significantly lower sparsity than DQN-RVM. This suggests that DQN-SBL performs better in the random-mapping feature space compared to the Gaussian kernel space. The sparsity achieved using the Gaussian kernel method is sensitive to the choice of hyperparameters. Therefore, as the number of training instances increases, SBELM is generally a more preferable choice over RVM due to its lower computational complexity, as discussed in [11].

*4) Comparison of Training time and complexity*

We conducted an evaluation of the training time, starting from the grid-search phase until the completion of the training process for each benchmark. The recorded training times are presented in Table IV. The training times for RVMs are very similar to each other within the same magnitude, while DQN-SBELM requires significantly more time compared to SBELM on certain datasets. However, both SBELM models can complete training within an hour, considering the 4 hidden nodes × 5 seeds = 20 grid-search trials, and the search itself can be completed within minutes. Figure 2 demonstrates that the accuracy of SBELM is not affected by the number of hidden nodes or seeds in these two models. In practical scenarios, the training time can be further reduced by narrowing down the scope of the grid search. Additionally, in most application scenarios, the main concern is the execution time with an accurate and sparse model, rather than the training time itself.

In addition to comparing the training time, we also compare the computational and storage complexity of our proposed models with the baseline models. Since most settings within RVMs and SBELMs remain the same, we primarily focus on the ratio of our proposed models compared to the baseline models in terms of complexity. We roughly define

$$computation_complexity_ratio(DQN\text{-}SBELM / SBELM) = \frac{O(L)}{O(L^3)} = 1/(max\ (L))^2$$

$$computation_complexity_ratio(DQN\text{-}RVM / RVM) = \frac{O(N)}{O(N^3)} = 1/N^2$$

$$storage_complexity_ratio(DQN\text{-}SBELM / SBELM) = \frac{O(L)}{O(L^2)} = 1/max\ (L)$$

$$storage_complexity_ratio(DQN\text{-}RVM / RVM) = \frac{O(N)}{O(N^2)} = 1/N$$

The complexity ratios are listed in Table V. The average computation and storage complexity ratio of DQN-RVM / RVM are on 1E-6 and 1E-3 respectively. This exponentially increases by the number of training data $N$. We can imagine that when $N$ is large enough, it will result to the memory-overflow issue and consume longer training time in the RVM. On the other hand, the average computation and storage complexity of DQN-SBELM / SBELM are about 1E-6 and 1E-3 respectively. As a larger number of hidden nodes than the feature dimension is required for the SBELMs model to enable better generalization. When the feature dimension increases, it will also result to the memory-issue and consume longer training time in SBELM model.

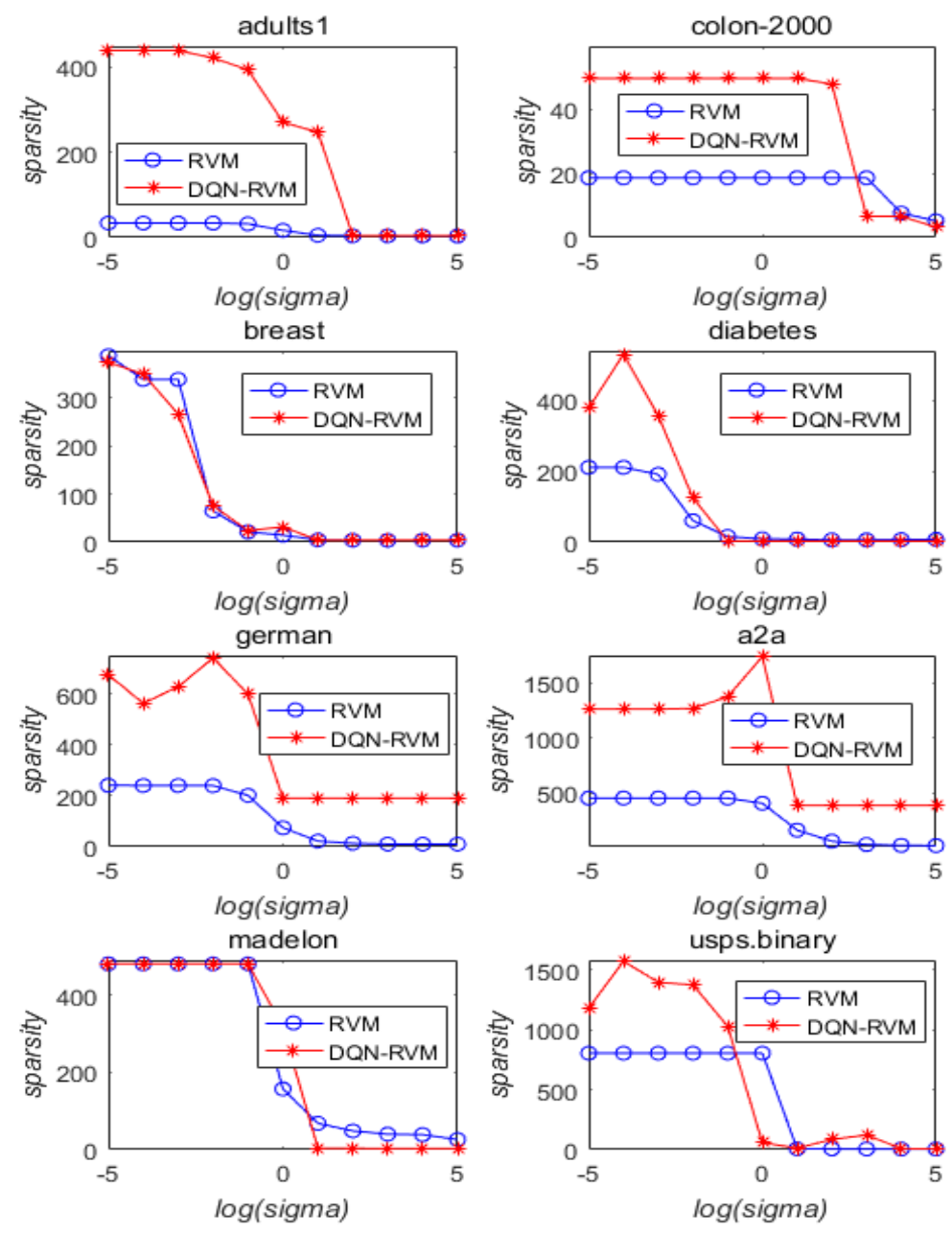

Fig. 3. Sparsity of RVMs under different hyperparameter ($log\ \sigma$)

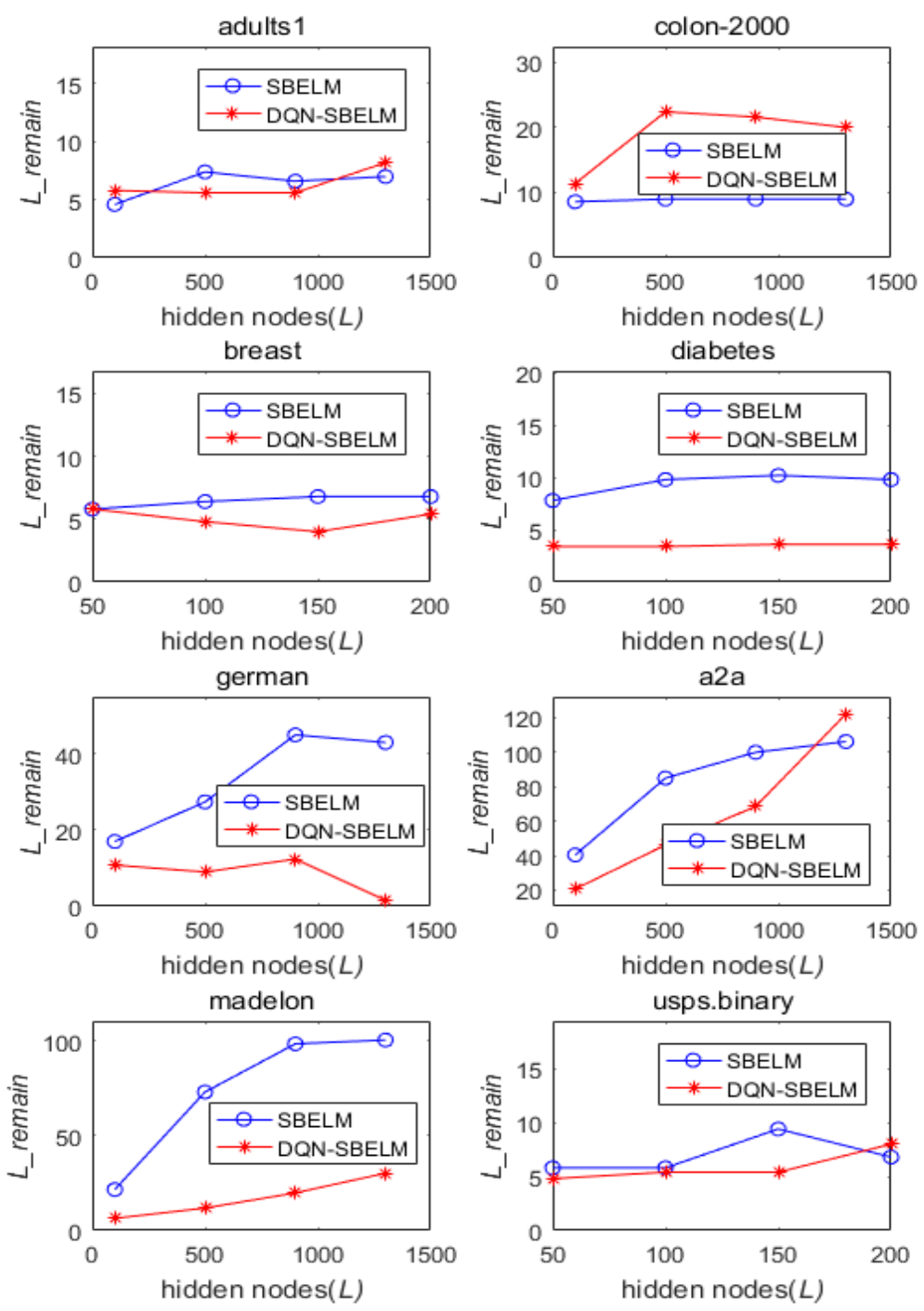

Fig. 4. Sparsity of SBELMs under different number of hidden nodes($L$).

In a nutshell, DQN-RVM achieves similar accuracy and training time compared to RVM. Although DQN-RVM exhibits higher sparsity in some benchmarks, the model size remains small and the computational complexity is $O(N)$ in contrast to the $O(N^3)$ complexity of RVM. This makes DQN-RVM scalable to large problems with a large training data size $N$. On the other hand, DQN-SBELM inherits the advantages of SBELM in terms of achieving a very sparse model and high generalization, albeit with slightly longer training time. The computational complexity of DQN-SBELM is $O(L)$ compared to $O(L^3)$ in SBELM. As the feature dimension increases, a larger L is required for competitive generalization, but this may lead to memory overflow in SBELM. DQN-SBELM, however, can avoid this issue. Importantly, both DQN-RVM and DQN-SBELM overcome the memory requirements of RVM and SBELM for large problems, aligning with the research objective of our proposed DQN-SBL.

### B. *DQN-SBL for Linear Classification*

#### 1) *Experimental Setup*

This section focuses on evaluating DQN-SBL on large-scale linear classifications with very high dimensional feature space where memory overflow may arise using SBL for such applications. Previous work on using SBL for linear classification is mainly on small problems in the literature. For our experiments, we selected four benchmarks: *news20, real-sim, rcv1.binary.train,* and *url_combined*, which are used for document classification. These datasets are randomly split into 75% for training data and 25% for test data.

To provide a comparison, we selected the $l_2 - norm$ logistic regression (LR-l2). LR-l2 is chosen because it shares a similar likelihood function with SBL, but it uses a unified regularization parameter λ to control the complexity of the weight vector w. In contrast, SBL replaces the unified $\lambda$ with a series of $\alpha_k$ $(k = 0, \ldots, M)$ to independently govern each component $w_k$. In DQN-SBL, each $\alpha_k$ is automatically determined from the training data, while the best $\lambda$ in LR-l2 is selected through exhaustive cross-validation.

TABLE V
COMPARISON OF ACCURACY

| Dataset | DQN-SBL | LR-l2 |
|---|---|---|
| *news20* | 96.85 | 97.06 |
| *real-sim* | 97.25 | 97.34 |
| *rcv1.binary.train* | 96.50 | 96.72 |
| *url_combined* | 98.54 | 97.89 |

#### 2) *Accuracy*

As SBL and its variants provide a trade-off between the sparsity and accuracy, its accuracy may mildly deteriorate to be underfitting with increasing iterations at **A-stage** in some problems. Therefore, in terms of best accuracy, we conduct the experiments without consideration of sparsity for DQN-SBL. An extra 5-fold cross-validation strategy is employed for DQN-SBL to seach the '*early-stopping*' iteration *iter* with best validation accuracy. After that, we train a new classifier terminated at the *iter* for predicting the test dataset. For the best hyper-parameter $\lambda$ in LR-l2, it is also achieved by a 5-fold cross-validation at scope $\lambda \in 2^{[-5,5]}$ with increasing step size by 2.

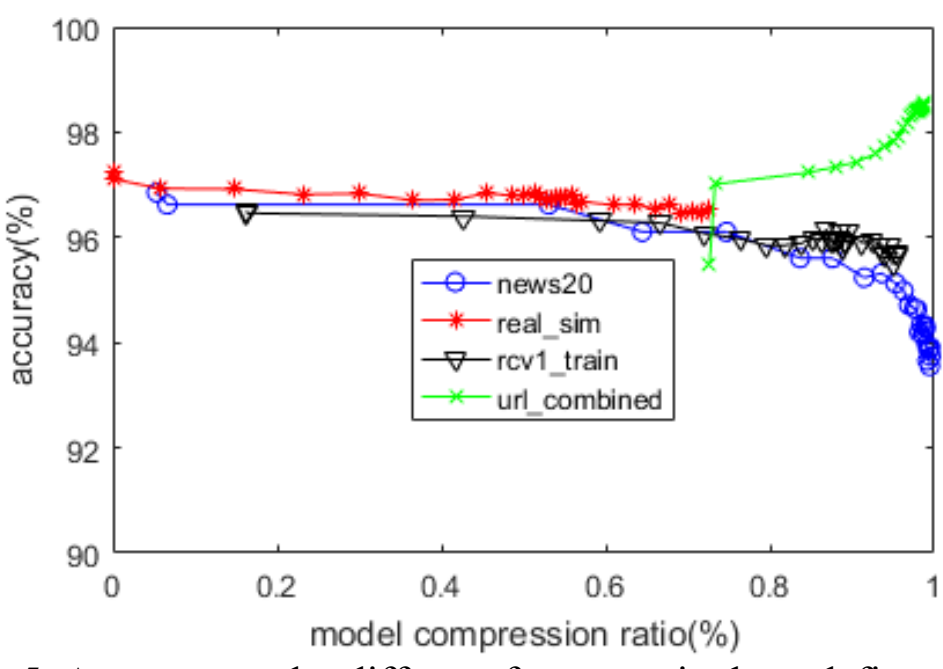

Fig. 5 Accuracy under different feature ratio, here define ratio=1-size(**w**!=0)/*M*.

Table V displays the accuracy results for both models. It is evident that the best accuracies achieved by the two models are very close. However, it is important to note that finding the best $\lambda$ for LR-l2 through exhaustive cross-validations can be burdensome for large-scale problems. This is particularly true when the data cannot be loaded into memory all at once, as it requires repeatedly loading data into memory during the search for the optimal $\lambda$. In contrast, DQN-SBL eliminates the need for hyperparameter selection as the $\boldsymbol{\alpha}$ values are automatically learned during the training phase.

#### 3) *Model Compression of DQN-SBL*

To verify the assumption that generalization does not sharply deteriorate with higher sparsity, we plot the evolution curves of accuracies on the four datasets at different sparsity levels in Fig. 5. In *news20*, *real-sim*, and *rcv1.binary.train* datasets, the accuracies mildly decrease as the sparsity increases. However, in the case of *url_combined* dataset, the remaining feature size decreases while the accuracy actually increases. This suggests that the relationship between sparsity and accuracy can vary depending on the dataset, and in some cases, higher sparsity can even lead to improved accuracy.

## V. DISCUSSION

The main complexity in SBL arises from calculating the covariance matrix $\boldsymbol{\Sigma}$ in the MAP function, where the $\boldsymbol{\Sigma}$ is essential for updating the regularization parameter $\boldsymbol{\alpha}$. In this paper, DQN-SBL is proposed as an alternative approach. It utilizes a diagonal quasi-Newton method for the **MAP** function, where the calculation of the covariance matrix is approximated by a diagonal matrix $\boldsymbol{B}$ and iteratively updated using the differences between $\mathbf{w}$ and the gradients $\nabla L(\mathbf{w})$. Once convergence is achieved, the diagonal elements of $\boldsymbol{B}$ are used to update the regularization parameter $\boldsymbol{\alpha}$ alternately. The storage and computational complexity of $\boldsymbol{B}$, as given in equation (13), is $O(M)$ in DQN-SBL ($O(N)$ in DQN-RVM, $O(L)$ in DQN-SBELM). The complexity of matrix inversion in SBL, which is $O(M^3)$, is reduced to $O(M)$ in DQN-SBL. This makes DQN-SBL particularly suitable for large-scale problems. The complexities are summarized in Table VI.

Strictly speaking, the datasets used to evaluate DQN-SBL in this paper may not be truly large-scale data that cannot be loaded into memory or processed by available computational resources. The experiments conducted in this paper aim to verify that DQN-SBL achieves similar accuracy and sparsity to SBL but with a linear complexity of $O(M)$, making it well-suited for scaling to large-scale problems. The experimental results described above confirm this conclusion.

TABLE VI
COMPLEXITY OF SBL VARIANTS

| Model | Computation complexity | Storage complexity |
|---|---|---|
| SBL | $O(M^3)$ | $O(M^2)$ |
| **DQB-SBL** | $O(M)$ | $O(M)$ |
| RVM | $O(N^3)$ | $O(N^2)$ |
| **DQN-RVM** | $O(N)$ | $O(N)$ |
| SBELM | $O(L^3)$ | $O(L^2)$ |
| **DQN-SBELM** | $O(L)$ | $O(L)$ |

*M*: feature dimension. *N*: number of training data. *L*: number of hidden nodes in SBELMs.

## VI. CONCLUSION

This paper extends the efficient learning technique SBL for large-scale learning by developing a diagonal Quasi-newton method (DQN-SBL) for its MAP function. The new method overcomes the limitation of SBL that requires a memory-intensive covariance matrix inversion with complexity $O(M^3)$ ($M$: feature size) in the MAP to alternately update the regularized parameters. In DQN-SBL, the matrix inversion is approximated by a positive diagonal matrix and iteratively updated using the previous **w** and gradients. In this way, the computational complexity and storage memory are both reduced to $O(M)$ and therefore DQN-SBL scales well to large-scale problems.

Based on DQN-SBL, DQN-RVM and DQN-SBELM are proposed in this paper. DQN-RVM is proposed to overcome the complexity $O(N^3)$ ($N$: training data size) of RVM for basis selection and achieves analogous accuracy and sparsity to RVM, but with complexity $O(N)$. Similarly, DQN-SBELM reduces the complexity of SBELM from $O(L^3)$ to $O(L)$( $L$: num of hidden nodes) for high dimensional problems. As a summary, DQN-SBL is very suitable for large-scale problems. The paper also demonstrates the advantage of DQN-SBL in large-scale linear classification problems, where it achieves competitive accuracy with an extremely small model size.